\documentclass{article}

\usepackage[preprint,nonatbib]{neurips_2024}

\usepackage[utf8]{inputenc} 
\usepackage[T1]{fontenc}    
\usepackage[pdftex]{graphicx}
\usepackage[numbers]{natbib}
\usepackage{overpic}
\usepackage{url}            
\usepackage{booktabs}       
\usepackage{amsfonts}       
\usepackage{nicefrac}       
\usepackage{microtype}      
\usepackage[dvipsnames,table,xcdraw]{xcolor}
\usepackage{amsmath}
\usepackage{amssymb}
\usepackage{pifont}
\usepackage{dsfont}
\usepackage{bm}
\usepackage{multirow}
\usepackage{algorithm}
\usepackage{algorithmicx}
\usepackage{algcompatible}
\usepackage{hyperref}       
\usepackage{tikz}
\usepackage{wrapfig} 
\usepackage{listings}

\usepackage[capitalize]{cleveref}
\crefname{section}{Sec.}{Secs.}
\Crefname{section}{Section}{Sections}
\Crefname{table}{Table}{Tables}
\crefname{table}{Tab.}{Tabs.}

\newcommand{\figref}[1]{Fig.~\ref{#1}}
\newcommand{\tabref}[1]{Tab.~\ref{#1}}
\newcommand{\eqnref}[1]{Eqn.~(\ref{#1})}
\newcommand{\secref}[1]{Sec.~\ref{#1}}

\newcommand{\best}[1]{{\color{red}{\bf{#1}}}}
\newcommand{\second}[1]{{\color{blue}{\underline{#1}}}}
\definecolor{myred}{RGB}{255, 0, 0}
\definecolor{mygreen}{RGB}{0, 178, 86}
\definecolor{myblue}{RGB}{4, 109, 190}
\definecolor{myflesh}{RGB}{242, 170, 132}
\definecolor{myyellow}{RGB}{255, 192, 0}
\definecolor{myhref}{RGB}{236, 107, 138}

\newcommand*\samethanks[1][\value{footnote}]{\footnotemark[#1]}

\newcommand{\framework}{LE3D}

\title{Lighting Every Darkness with 3DGS:
Fast Training and Real-Time Rendering for HDR View Synthesis}

\author{%
  {\textbf{Xin Jin}}$^{1,2}$\thanks{\small{Equal Contribution. This project is done during Xin Jin's internship at MEGVII Technology.}}\quad
  {\textbf{Pengyi Jiao}}$^1$\samethanks\quad
  {\textbf{Zheng-Peng Duan}}$^1$\quad
  {\textbf{Xingchao Yang}}$^2$\quad\\
  {\textbf{Chun-Le Guo}}$^1$\thanks{\small{C. L. Guo and B. Ren ({\tt{\{guochunle,rb\}@nankai.edu.cn}}) are corresponding authors.}}\quad
  {\textbf{Bo Ren}}$^1$\samethanks\quad    
  {\textbf{Chongyi Li}}$^1$\quad\\
  {\small{$^1$ VCIP, CS, Nankai University}}\quad
  {\small{$^2$ MEGVII Technology}}\\
  {\tt{\textcolor{myhref}{\scriptsize{\href{https://srameo.github.io/projects/le3d}{https://srameo.github.io/projects/le3d}}}}}
}

\begin{document}

\maketitle

\vspace{-9mm}
\begin{figure*}[h]
  \centering
  \begin{overpic}[width=\textwidth]{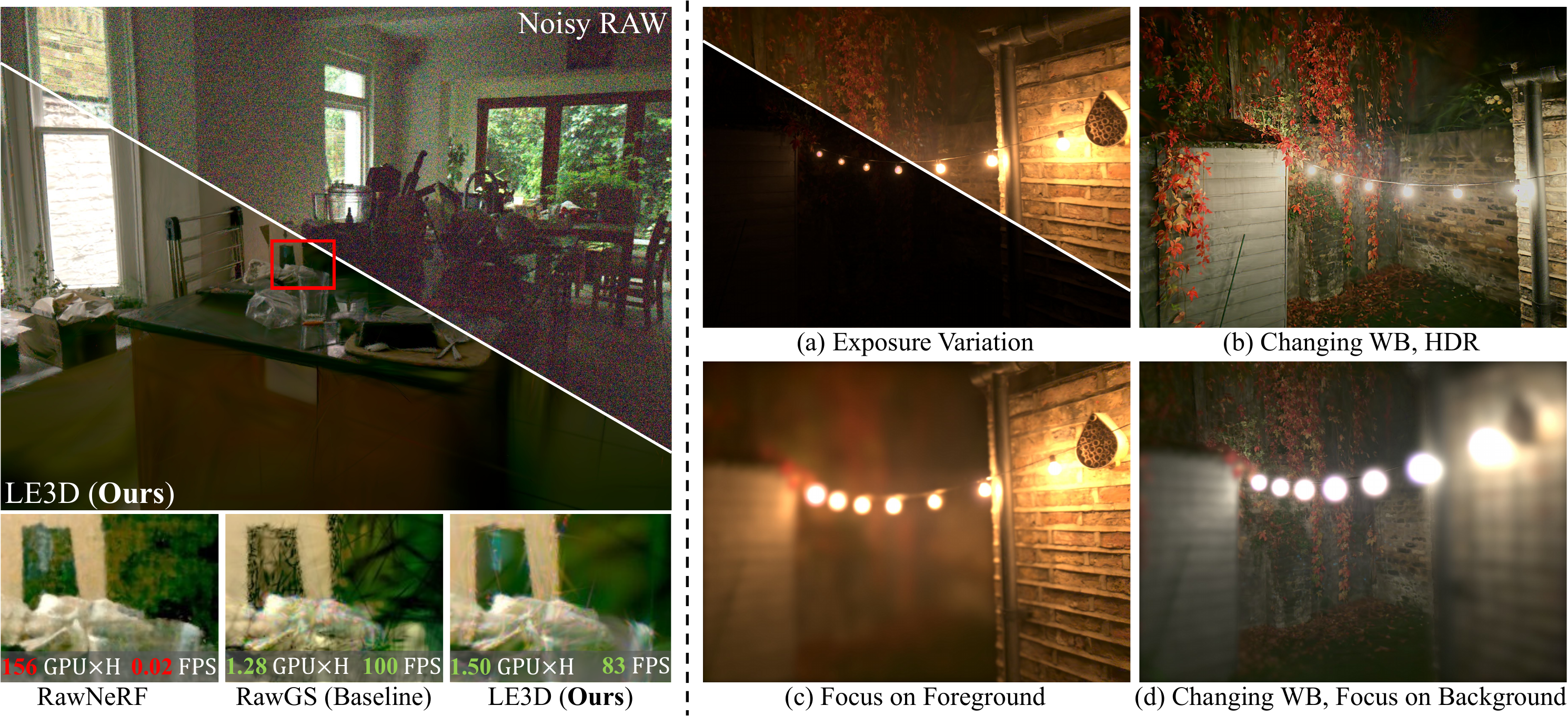}
    \put(9.8,0.7){\tiny{\cite{mildenhall2022nerf}}}
  \end{overpic}
  \vspace{-7mm}
  \caption{
    \framework~reconstructs a 3DGS representation of a scene from a set of multi-view noisy RAW images.
    As shown on the left, \framework~features \textit{fast training and real-time rendering} capabilities compared to RawNeRF~\cite{mildenhall2022nerf}. 
    Moreover, compared to RawGS (a 3DGS~\cite{kerbl20233d} we trained with RawNeRF's strategy), \framework~demonstrates superior noise resistance and 
    the ability to represent HDR linear colors.
    The right side highlights the variety of \textit{real-time downstream tasks} \framework~can perform, including (a) exposure variation, (b, d) changing White Balance (WB), (b) HDR rendering, and (c, d) refocus.
  }
  \label{fig:teaser}
\end{figure*}
\vspace{-1mm}
\vspace{-1mm}
\begin{abstract}
\vspace{-3mm}
Volumetric rendering based methods, like NeRF, excel in HDR view synthesis from RAW images, especially for nighttime scenes. 
While, they suffer from long training times and cannot perform real-time rendering due to dense sampling requirements.
The advent of 3D Gaussian Splatting (3DGS) enables real-time rendering and faster training. 
However, implementing RAW image-based view synthesis directly using 3DGS is challenging due to its inherent drawbacks:
1) in nighttime scenes, extremely low SNR leads to poor structure-from-motion (SfM) estimation in distant views;
2) the limited representation capacity of spherical harmonics (SH) function is unsuitable for RAW linear color space; and
3) inaccurate scene structure hampers downstream tasks such as refocusing.
To address these issues, we propose \textbf{LE3D} (\textbf{L}ighting \textbf{E}very darkness with \textbf{3D}GS).
Our method proposes Cone Scatter Initialization to enrich the estimation of SfM, 
and replaces SH with a Color MLP to represent the RAW linear color space.
Additionally, we introduce depth distortion and near-far regularizations to improve the accuracy of scene structure for downstream tasks.
These designs enable \framework~to perform real-time novel view synthesis, HDR rendering, refocusing, and tone-mapping changes. 
Compared to previous volumetric rendering based methods, \framework~reduces training time to 1\% and improves rendering speed by up to 4,000 times for 2K resolution images in terms of FPS. 
Code and viewer can be found in \href{https://github.com/Srameo/LE3D}{https://github.com/Srameo/LE3D}.
\end{abstract}
\section{Introduction}
\label{sec:intro}

Since the advent of Neural Radiance Fields (NeRF)~\cite{mildenhall2020nerf}, novel view synthesis (NVS) has entered a period of vigorous development, 
thereby advancing the progress of related applications in augmented and virtual reality (AR/VR).
Existing NVS technologies predominantly utilize multiple well-exposed 
and noise-free low dynamic range (LDR) RGB images as inputs to reconstruct 3D scenes.
This significantly impacts the capability to capture high-quality images in environments with low light or high contrast, 
such as nighttime settings or areas with stark lighting differences. 
These scenarios typically necessitate the use of high dynamic range (HDR) scene reconstruction techniques.

Existing HDR scene reconstruction techniques primarily fall into two categories and all are based on volumetric rendering: 
1) using multiple-exposure LDR RGB images for supervised training~\cite{huang2022hdr}, and 
2) conducting training directly on noisy RAW data~\cite{mildenhall2022nerf}.
The first type of method is typically highly effective in well-lit scenes. 
However, in nighttime scenarios, its reconstruction performance is constrained due to the impact of the limited dynamic range in sRGB data~\cite{chen2018learning}.
While RAW data preserves more details in nighttime scenes, it is also more susceptible to noise.
Therefore, RawNeRF~\cite{mildenhall2022nerf} is proposed to address the issue of vanilla NeRF's lack of noise resistance.
However, RawNeRF suffers from prolonged training times and an inability to render in real-time (a common drawback of volumetric rendering-based methods).
This significantly limits the application of current scene reconstruction techniques and HDR view synthesis.
Enabling real-time rendering for HDR view synthesis is a key step towards bringing computational photography to 3D world.

Recently, 3D Gaussian Splatting (3DGS) has garnered significant attention based on
its powerful capabilities in real-time rendering and photorealistic reconstruction.
3DGS utilizes Structure from Motion (SfM)~\cite{schoenberger2016sfm} for initialization and 
employs a set of 3D gaussian primitives to represent the scene.
Each gaussian represents direction-aware colors using spherical harmonics (SH) functions and
can be updated in terms of color, position, scale, rotation, and opacities through gradient descent optimization.
Although 3DGS demonstrates its strong capabilities in reconstruction and real-time rendering, 
it is not suitable for direct training using nighttime RAW data.
This is primarily due to
1) the SfM estimations based on nighttime images are often inaccurate, leading to blurred distant views 
or the potential emergence of floaters; 
2) the SH does not adequately represent the HDR color information of RAW images due to its limited representation capacity; and
3) the finally reconstructed structure, such as depth map, 
is suboptimal, leading to unsatisfactory performance in downstream tasks like refocus.

To make 3DGS suitable for HDR scene reconstruction, we introduce \framework~that stands for {\textbf{L}}ighting {\textbf{E}}very darkness with {\textbf{3D}}-GS, 
addressing the three aforementioned issues.
First, to address the issue of inaccurate SfM distant view estimation in low-light scenes, 
we proposed Cone Scatter Initialization to enrich the COLMAP-estimated SfM. 
It performs random point sampling within a cone using the estimated camera poses.
Second, instead of the SH, we use a tiny MLP to represent the color in RAW linear color space.
To better initialize the color of each gaussian, different color biases are used for various gaussian primitives.
Thirdly,
we propose to use depth distortion and near-far regularization to achieve better scene structure for downstream task like refocusing. 
As shown in~\figref{fig:teaser} (left), our \framework~can perform real-time rendering with only 
about 1.5 GPU hours (99$\%$ less than RawNeRF~\cite{mildenhall2022nerf}) of training time and at a rate of 100 FPS 
(about 4000$\times$ faster than RawNeRF~\cite{mildenhall2022nerf}) with comparable quality.
Additionally, it is capable of supporting downstream tasks such as HDR rendering, refocusing, and exposure variation, as shown in~\figref{fig:teaser} (right).

In summary, we make the following contributions:
\begin{itemize}
    \item We propose \framework, which can reconstruct HDR 3D scenes from noisy raw images and perform real-time rendering. Compared with the NeRF-based methods, \framework~requires only 1\% of the training time and has 4000$\times$ render speed.
    \item To address the deficiencies in color representation by vanilla 3DGS and the inadequacies of SfM estimation in nighttime scenes, 
          we leverage a Color MLP with primitive-aware bias, and introduce Cone Scatter Initialization to enrich the point cloud initialized by COLMAP.
    \item To improve the scene structure in the final results for achieving better downstream task performance, 
          we introduce depth distortion and near-far regularizations. 
\end{itemize}

\section{Related work}

\paragraph{Novel view synthesis}
Since the emergence of NeRF~\cite{mildenhall2020nerf}, NVS has gained significant advancement. 
NeRF employs an MLP to represent both the geometry of the scene and the view-dependent color. 
It utilizes the differentiable volume rendering~\cite{levoy1990efficient}, 
thereby enabling gradient-descent training through a multi-view set of 2D images.
Subsequent variants of NeRF~\cite{barron2021mip,barron2022mip,hu2023tri} have extended NeRF's capabilities with anti-aliasing features.
To overcome the deficiencies of vanilla NeRF in geometry reconstruction, 
strategies such as depth supervision~\cite{deng2022depth,dadon2023ddnerf} and distortion loss~\cite{barron2022mip} have been introduced into NeRF.
Some methods~\cite{fridovich2022plenoxels,chen2022tensorf,muller2022instant,chen2023neurbf} have conducted experiments with feature-grid based approaches to enhance the training and rendering speeds of NeRF. 
Although these methods have achieved relatively promising results in novel view synthesis, 
the training and rendering speeds are still significant bottlenecks. 
This issue is primarily due to the dense sampling inherently required by volume rendering.

Recently, the advent of 3D Gaussian Splatting~\cite{kerbl20233d} has marked a significant advancement in real-time NVS methods. 
3DGS represents a scene using a collection of 3D gaussian primitives, each endowed with distinct attributes. 
Some subsequent works have added anti-aliasing capabilities to 3DGS representations~\cite{yu2024mip,song2024sa,liang2024analytic}; 
others have enhanced 3DGS representation capabilities through supervision in the frequency domain~\cite{zhang2024fregs}. 
DNGaussian~\cite{li2024dngaussian} proposed a depth-regularized framework to optimize sparse-view 3DGS, and other works also relying on depth supervision~\cite{chung2023depth,kung2024sad}. 
Additionally, some works~\cite{li2023spacetime,wu20234d,luiten2023dynamic} have focused on applying 3DGS to dynamic scene representation.
However, these methods only accept LDR sRGB data for training, and thus cannot reconstruct the scene's HDR radiance. 
This limitation suggests they cannot perform downstream tasks such as HDR tone mapping and exposure variation. 
In contrast, \framework~is specifically designed to reconstruct the HDR representation of scenes from noisy RAW images.

\paragraph{HDR view synthesis and its applications}
HDR typically refers to a concept in computational photography that focuses on preserving as much dynamic range as possible to facilitate more post-processing options~\cite{debevec1997recovering,eilertsen2017hdr,liu2020single,kalantari2017deep,hasinoff2016burst}. 
The existing HDR view synthesis techniques bear a striking resemblance to the two main approaches in 2D image HDR synthesis:
1) Direct use of multiple-exposure LDR images to compute the camera response function (CRF) and synthesize an HDR image~\cite{debevec1997recovering}. 
   This corresponds to HDR-NeRF~\cite{huang2022hdr} which employs an MLP to learn the CRF.
2) Acquisition of noise-free underexposed RAW images, utilizing the characteristics of the linear color space in RAW to manually simulate multiple-exposure images, and synthesize an HDR image. 
   This corresponds to RawNeRF~\cite{mildenhall2022nerf}, which learns a NeRF representation of RAW linear color space with noisy RAW images to perform both denoising and NVS.
Although both methods achieve impressive visual results, the dense sampling required by volume rendering still poses a bottleneck for both training time and rendering efficiency.
\framework~follows the same technical approach as RawNeRF, reconstructing scene representations from noisy RAW images. 
This means that \framework~does not necessarily require training data with multiple exposures, significantly broadening its range of applications.
However, a key distinction of \framework~is its use of differentiable rasterization techniques~\cite{kerbl20233d,kopanas2021point,sun2022direct}, 
which enable \textit{fast training and real-time rendering}.
Based on a 3DGS-like representation of the reconstructed scene, \framework~can perform real-time HDR view synthesis.
This is a novel attempt to introduce computational photography into 3D world, as it enables \textit{real-time reframing and post-processing} (changing white balance, HDR rendering, etc. as shown in~\figref{fig:teaser}).
\section{Preliminaries}

\paragraph{3D Gaussian Splatting}
3D Gaussian Splatting renders detailed scenes by computing the color and depth of pixels through the blending of many 3D gaussian primitives. Each gaussian is defined by its center in 3D space $\mu_i \in \mathbb{R}^3$, a scaling factor $s_i \in \mathbb{R}^3$, a rotation quaternion $q_i \in \mathbb{R}^4$, and additional attributes such as opacity $o_i$ and color features $f_i$. The basis function of a gaussian primitive is given by \eqnref{eq:gau_pri} that incorporates the covariance matrix derived from the scaling and rotation parameters.
\begin{equation}
    G(x) = \exp(-\frac{1}{2}(x-\mu)^T\Sigma^{-1}(x-\mu)).
    \label{eq:gau_pri}
\end{equation}
During rendering, the color of a pixel is determined by blending the contributions of multiple gaussians that overlap the pixel's location. This process involves decoding the color features $f_i$ to color $c_i$ by the SH, and calculating $\alpha_i$ of each primitive by multiplied its opacity $o_i$ with its projected 2D gaussian $G_i^{2D}$ on the image plane. Unlike traditional ray sampling strategies, 3D Gaussian Splatting employs an optimized rasterizer to gather the relevant gaussians for rendering. Specifically, the color $C$ is computed by blending $N$ ordered gaussians overlapping the pixel:

\begin{equation}
    C = \sum_{i \in N} c_i \cdot \alpha_i \prod_{j=1}^{i-1}(1-\alpha_j), \mbox{where}\; \alpha_i = G_i^{2D} o_i.
\end{equation}

\paragraph{HDR view synthesis with noisy RAW images}
RawNeRF~\cite{mildenhall2022nerf}, as a powerful extension of NeRF, specifically addresses the challenge of high dynamic range (HDR) view synthesis with noisy images. Different from LDR images, the dynamic range in HDR images can span several orders of magnitude between bright and dark regions, resulting in the NeRF's standard L2 loss being inadequate. To address this challenge, RawNeRF introduces a weighted L2 loss that enhances supervision in the dark regions. RawNeRF applies gradient supervision on tone curve $\psi=\log(y+\epsilon)$ with $\epsilon=10^{-3}$, and uses $\psi'=(y+\epsilon)^{-1}$ as the weighting term on the L2 loss between rendered color $\hat{y}_i$ and noisy reference color $y_i$. Applying the stop-gradient function $sg(\cdot)$ on $y$, the final loss can be expressed as:
\begin{equation}
    L_{\psi}(\hat{y},y)=\sum_i(\frac{\hat{y}_i-y_i}{sg(\hat{y}_i)+\epsilon})^2.
    \label{eq:rawnerf_pri}
\end{equation}

Moreover, RawNeRF employs a variable exposure training method to take advantage of images with varying shutter speeds. The method scales the output color in linear RGB space by a learned factor $\beta_{t_i}^c$ for each color channel $c$ with each unique shutter speed $t_i$. In particular, the $c$-th channel of the output color $\hat{y}_i^c$ will be mapped into $\min(\hat{y}_i^c \cdot t_i \cdot \beta_{t_i}^c, 1)$ as the final output for rendering.
\section{Proposed method}

\begin{figure*}[t]
  \centering
  \begin{overpic}[width=\textwidth]{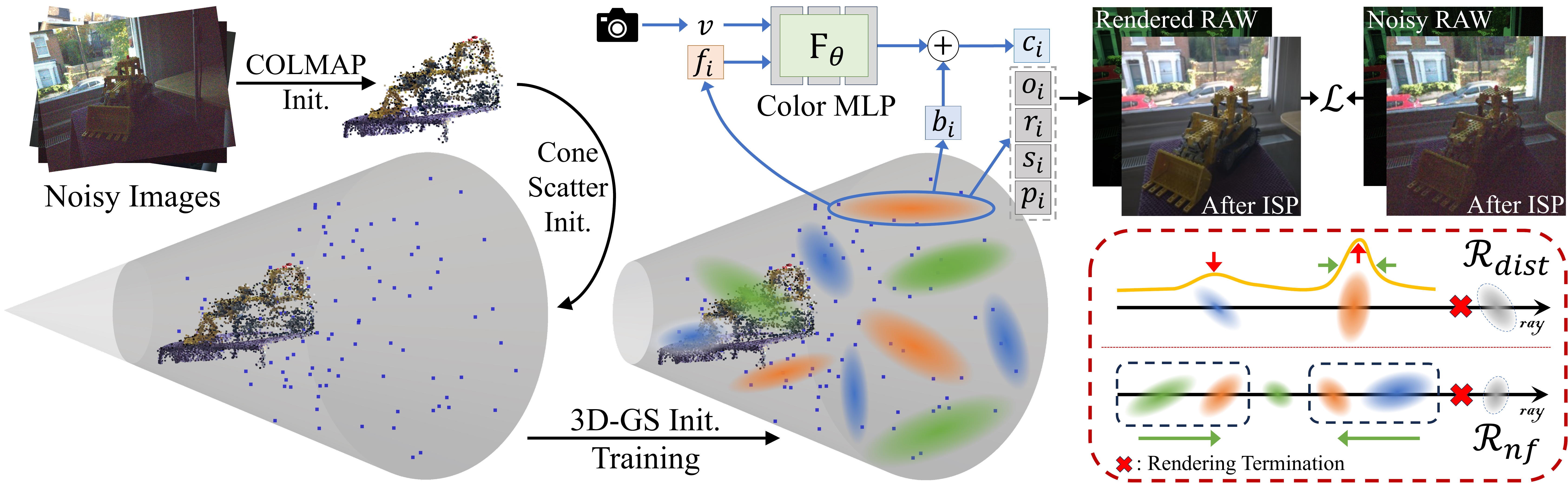}
    \put(8,3){\large{$\mathcal{F}$}}
    \put(72,13){{\textcolor{myyellow}{$\mathcal{H}$}}}
  \end{overpic}
  \caption{
    Pipeline of our proposed \framework. 
    1) Using COLMAP to obtain the initial point cloud and camera poses. 
    2) Employing Cone Scatter Initialization to enrich the point clouds of distant scenes. 
    3) The standard 3DGS training, where we replace the original SH with our tiny Color MLP to represent the RAW linear color space. 
    4) We use RawNeRF's weighted L2 loss $\mathcal{L}$ (\eqnref{eq:rawnerf_pri}) as image-level supervision,
    and our proposed $\mathcal{R}_{dist}$ (\eqnref{eq:depth_dist}) as well as $\mathcal{R}_{nf}$ (\eqnref{eq:near_far}) as scene structure regularizations. 
    In this context, $f_i$, $b_i$, and $c_i$ respectively represent the color feature, bias, and final rendered color of each gaussian $i$.
    Similarly, $o_i$, $r_i$, $s_i$, and $p_i$ denote the opacity, rotation, scale, and position of them.
  }
  \label{fig:pipeline}
\end{figure*}

The pipeline of our~\framework~is shown in~\figref{fig:pipeline}.
Our main motivations and solutions are as follows:
1) To address the issue of COLMAP's inadequacy in capturing distant scenes during nighttime, we utilize the proposed Cone Scatter Initialization to enrich the point cloud obtained from COLMAP.
2) Experiments show that the original SH in 3DGS is inadequate for representing the RAW linear color space (as shown in \figref{fig:ablation} (e) and \figref{fig:supp_quant_ablation}). Therefore, we replace it with a tiny color MLP.
3) To enhance the scene structure and optimize the performance of downstream tasks, we propose the depth distortion $\mathcal{R}_{dist}$ and near-far $\mathcal{R}_{nf}$ regularizations.

\subsection{Improvements to the vanilla 3DGS representation}

Directly applying 3DGS on noisy RAW image set faces aforementioned two challenges, lack of distant points and inadequate representation of RAW linear color space. To address them, we propose the following improvements to the vanilla 3DGS representation.

\vspace{-2mm}
\paragraph{Cone Scatter Initialization}
To enrich the COLMAP-initialized point cloud $\mathcal{S}=\{\mathbf{s}_i\}$ with distant scenes, we estimate the position and orientation of all cameras. 
Based on this, we randomly scatter points within a predefined viewing frustum $\mathcal{F}$. 
To define $\mathcal{F}$, we need to determine:
1) The viewpoint $\mathbf{p}$; 2) The viewing direction $\vec{\mathbf{n}}$; 3) The field of view $\Theta$; and 4) The near and far planes, $z = z_n$ and $z = z_f$, respectively.
For forward-facing scenes, the viewing direction can be easily determined by averaging the orientations of all cameras, represented as $\vec{\mathbf{n}}=\mathrm{avg}\{\vec{\mathbf{n}}^\mathrm{c}_i\}$.
To encompass all visible areas in space from the training viewpoints, we use the maximum value of FOV from all cameras, denoted as $\Theta=\max\{\theta^\mathrm{c}_i\}$.
Additionally, $\mathcal{F}$ needs to include all the camera origins $\{\mathbf{p}^\mathrm{c}_i\}$ to ensure complete coverage of the scene from all perspectives.
It means that $\mathcal{F}$ should encompass a circle with its center at $\overline{\mathbf{p}}^\mathrm{c}=\mathrm{avg}\{\mathbf{p}^\mathrm{c}_i\}$, 
radius $r=\max{\{\left\|\mathbf{p}^\mathrm{c}_i-\overline{\mathbf{p}}^\mathrm{c}\right\|_2\}}$, and perpendicular to $\vec{\mathbf{n}}$.
Therefore, we can establish $\mathcal{F}$:
\begin{equation}
\begin{split}
  &\mathbf{p}=\overline{\mathbf{p}}^\mathrm{c}-\frac{r}{\tan(\Theta/2)}\cdot\frac{\vec{\mathbf{n}}}{\left\|\vec{\mathbf{n}}\right\|_2},\;\;
  \vec{\mathbf{n}}=\mathrm{avg}\{\vec{\mathbf{n}}^\mathrm{c}_i\},\;\;
  \Theta=\max\{\theta^\mathrm{c}_i\},\\
  &z_n=\min\{\left\|\mathbf{s}_i-\mathbf{p}\right\|_2\},\;\;
  z_f=\lambda_\mathcal{F}\cdot\max\{\left\|\mathbf{s}_i-\mathbf{p}\right\|_2\}.
  \label{eq:view_frustum}
\end{split}
\end{equation}
For near $z_n$ and far $z_f$, we use the distance from the nearest and 
$\lambda_\mathcal{F}$ times the distance from farthest points in the COLMAP-initialized point cloud $\mathcal{S}$
to $\mathbf{p}$ to represent them, respectively. 
Subsequently, we randomly scatter points within our viewing frustum 
$\mathcal{F}=\{\mathbf{p}, \vec{\mathbf{n}}, \Theta, z_n, z_f\}$
to obtain our enriched point cloud $\mathcal{S}^\prime=\mathcal{S} \cup \mathcal{S}^\mathcal{F}$, where $\mathcal{S}^\mathcal{F}$ is the scattered point set.
Then $\mathcal{S}^\prime$ is used for initialization of the gaussians, instead of $\mathcal{S}$.

\vspace{-2mm}
\paragraph{Color MLP with primitive-aware bias}
To address the issue that SH could not adequately represent the RAW linear color space, 
we replace it with a tiny color MLP $\mathbf{F}_\theta$.
Each gaussian is initialized with a random color feature $f_i$ and a color bias $b_i$.
To initialize $b_i$, 
we project each $\mathbf{s}_i^\prime\in \mathcal{S}^\prime$ onto every training image, obtaining a set of all pixels $\{c_\mathrm{pix}\}_i$ for each point $i$,. 
The color feature $f_i$ is concatented with the camera pose $v$, and then it is feeded into the tiny color MLP $\mathbf{F}_\theta$ to obtain the view dependent color.
Since the HDR color space theoretically has no upper bound on color values,
we use the exponent function as the activation function for $\mathbf{F}_\theta$ to simulate it.
The final color $c_i$ is:
\begin{equation}
  c_i=\exp\left(\mathbf{F}_\theta(f_i,v)+b_i\right), \mbox{where } b^{(0)}_i = \log(\mathrm{avg}(\{c_\mathrm{pix}\}_i)), f^{(0)}_i\gets \mathcal{N}(0, \sigma_f).
\end{equation} 
where $f^{(0)}_i$ is sampled from a gaussian distribution $\mathcal{N}(0, \sigma_f)$ and $b^{(0)}_i$ is setted by the log value of the average of $\{c_\mathrm{pix}\}_i$.
This initialization makes $c_i^{(0)}$ close to $\mathrm{avg}\{c_\mathrm{pix}\}_i$.
Both $f_i$ and $b_i$ are learnable parameters and during cloning and splitting, 
they are copied and assigned to new gaussians.

\subsection{Depth distortion \& near-far regularizations}
\label{sec:method_regs}
Scene structure is crucially important for the downstream applications of our framework, 
particularly the tasks such as refocusing.
Therefore, we propose depth distortion and near-far regularizations to enhance the ability of 3DGS for optimizing scene structure. 
Borrow from NeRF-based methods~\cite{barron2022mip}, we use depth map and weight map to regularize the scene structure.

\vspace{-2mm}
\paragraph{Depth and weight map rendering} 
Recently, several 3DGS-based works~\cite{li2024dngaussian,chung2023depth} employ some form of supervision on depth. 
Also, depth maps are crucial for downstream tasks such as refocus (\secref{sec:applications}), mech extraction~\cite{guedon2023sugar} and relighting~\cite{gao2023relightable,zhang2024darkgs}.
They are achieved by obtaining the rendered average depth map $d$ in the following manner:
\begin{equation}
d=\frac{\sum_{i} z^\mathrm{c}_i \omega_i}{\sum_{i} \omega_i},\;\mbox{where}\;
[x^\mathrm{c}_i, y^\mathrm{c}_i, z^\mathrm{c}_i]^T= W[x_i, y_i, z_i]^T+t,\,\mbox{and}\;
\omega_i = \alpha_i \prod_{j=1}^{i-1}(1-\alpha_j).
\end{equation}
where $d$ denotes the depth map, $\omega_i$ represents the blending weight of the $i$-th gaussian,
$[x_i, y_i, z_i]^T$ and $[x^\mathrm{c}_i, y^\mathrm{c}_i, z^\mathrm{c}_i]^T$ represent the position in the world and the camera coordinate system, respectively,
and $[W, t]$ corresponds to the camera extrinsics.
The pixel values in the Weight Map each describe a histogram $\mathcal{H}$ of the distribution on the ray passing through this pixel.
Similar to Mip-NeRF 360~\cite{barron2022mip}, we can optimize the scene structure by 
constraining the gaussian primitives on each ray to be more concentrated.
To obtain the Weight Map, we first need to determine the distances to the nearest and farthest gaussian primitives 
from the current camera pose $p^\mathrm{c}$, represented as $z_n^\mathrm{c}, z_f^\mathrm{c}$, respectively.
Subsequently, we transform the interval $[z_n^\mathrm{c}, z_f^\mathrm{c})$ to obtain $K$ intersections, 
where the $k$-th intersection is denoted as $[t_k, t_{k+1})$.
Thus, the $k$-th value in the histogram $\mathcal{H}(k)$ can be obtained through rendering in the following manner:
\begin{equation}
\mathcal{H}(k)=\sum_{i} \mathds{1}{(z^\mathrm{c}_i, k)} \omega_i,\;
\mbox{where}\;
\mathds{1}{(z^\mathrm{c}_i, k)} = 
\begin{cases} 
1 & \text{if } z^\mathrm{c}_i \in [t_k, t_{k+1}) \\
0 & \text{else}
\end{cases}.
\label{eq:weight_distortion_map}
\end{equation}
Rendering $\mathcal{H}$ is essential, as it is effective not only in regularization 
but also plays a role in the refocusing application. 

\paragraph{Proposed regularizations}
Inspired by Mip-NeRF 360~\cite{barron2022mip}, we proposed similar depth distortion regularization $\mathcal{R}_{dist}$ to 
concentrate the gaussians on each ray:
\begin{equation}
\mathcal{R}_{dist}=
\sum^{K}_{u,v}\mathcal{H}(u)\mathcal{H}(v)
\left|\frac{t_{u}+t_{u+1}}{2}-\frac{t_{v}+t_{v+1}}{2}\right|.
\label{eq:depth_dist}
\end{equation}
$\mathcal{R}_{dist}$ constrains the weights along the entire ray to either approach zero or be concentrated at the same intersection.
However, in unbounded scenes of the real world, the distances $({z_f^\mathrm{c}-z_n^\mathrm{c}})/{K}$ between each intersection are vast. 
Forcibly increasing the size of $K$ to reduce the length of each intersection also significantly increases the computational burden.
This means that our $\mathcal{R}_{dist}$ can only provide a relatively coarse supervision for the gaussians on each ray, 
primarily by constraining them as much as possible within the same intersection.

To further constrain the concentration of gaussians, we propose the Near-Far Regularization $\mathcal{R}_{nf}$.
$\mathcal{R}_{nf}$ enhances the optimization of scene structure by narrowing the distance between the weighted depth of the nearest and farthest $M$ gaussians
on each ray, where the farthest refers to the last $M$ gaussians when the blending weight approaches $1$.
First, we extract two subsets of gaussians, $\mathbf{N}$ and $\mathbf{F}$, 
which respectively contain the nearest and farthest $M$ gaussians on each ray.
Subsequently, we render the depth maps for both subsets ($d^\mathbf{N}$, $d^\mathbf{F}$), as well as the final blending weight map ($T^\mathbf{N}$, $T^\mathbf{F}$). The blending weight map $T$ is the sum of each $\omega_i$. 
And here comes $\mathcal{R}_{nf}$:
\begin{equation}
\mathcal{R}_{nf} = T^\mathbf{N}\cdot T^\mathbf{F}\cdot\left|d^\mathbf{N} - d^\mathbf{F}\right|.
\label{eq:near_far}
\end{equation}
It not only can prune the gaussians at the front or back of each ray through opacity supervision when there is a significant disparity between them 
(relying on the $T^\mathbf{N}\cdot T^\mathbf{F}$ term).
Compared to $\mathcal{R}_{dist}$, $\mathcal{R}_{nf}$ can also supervise the position of the first and last $M$ gaussians on each ray to be as close as possible 
(relying on the $\left|d^\mathbf{N} - d^\mathbf{F}\right|$ term).
Besides the weighted L2 loss $\mathcal{L}$ and proposed regularizations $\mathcal{R}_{dist}$ and $\mathcal{R}_{nf}$, 
we also introduce constraints on the final blending weights $T$.
Given that the \framework~is tested in real-world scenarios, $T$ should ideally approach $1$, 
meaning that all pixels should be rendered. Thus, we propose $\mathcal{R}_T=-\log(T+\epsilon)$
to penalize the pixels where $T$ is less than 1.
\section{Experiments}

\subsection{Implementation details}
\label{sec:implementation}

\paragraph{Loss functions and regularizations}
In our implementation, the final loss function is:
\begin{equation}
L=\mathcal{L}+\lambda_T\mathcal{R}_T+\lambda_{dist}\mathcal{R}_{dist}+\lambda_{nf}\mathcal{R}_{nf},
\end{equation}
where $\mathcal{L}$ is the weighted L2 loss, and $\mathcal{R}_{T}$, $\mathcal{R}_{dist}$, and $\mathcal{R}_{nf}$ are the proposed 
T, depth distortion, and near-far regularizations, respectively.

\paragraph{Optimization}
We set $\lambda_\mathcal{F}$ to $10$ to enrich the COLMAP-initialized point cloud in distant views.
$\lambda_T, \lambda_{dist}, \lambda_{nf}$ in the loss function are set to $0.01, 0.1, 0.01$ respectively.
For our color MLP $\mathbf{F}_{\theta}$, we use the Adam optimizer with an initial learning rate of $1.0e-4$.
The initial learning rates for color features and biases for each gaussians are set to $2.0e-3$ and $1.0e-4$, respectively.
The learning rates for all three decrease according to a cosine decay strategy to a final learning rate of $1.0e-5$.
Besides the color MLP, primitive-aware color bias, and the color features for each gaussians, other settings are the same as those of 3DGS~\cite{kerbl20233d}.
For scenes captured with multiple exposures, we employ the same multiple exposure training strategy as RawNeRF~\cite{mildenhall2022nerf}.

\subsection{Datasets and comparisons}

\vspace{-1mm}
\paragraph{Datasets}
We evaluated \framework's performance on the benchmark dataset of RawNeRF. 
It includes fourteen scenes for qualitative testing and three test scenes for quantitative testing.
The three test scenes, each contains 101 noisy images and a clean reference image merged from stabilized long exposures.
However, the training data are captured with short exposures, leading to exposure inconsistencies. 
Therefore, we apply the same affine alignment operation as RawNeRF before testing (detailed in~\secref{sec:supp_affine_alignment} of the supplementary materials).
All images are $4032\times 3024$ Bayer RAW images captured by an iPhone X, saved in DNG format.

\vspace{-1mm}
\paragraph{Baseline and comparative methods}
We compare two categories of methods, 3DGS-based methods and NeRF-based methods.
The baseline we compare against is RawGS which uses vanilla 3DGS for scene representation and employs the weighted L2 loss and multiple exposure training proposed in RawNeRF~\cite{mildenhall2022nerf}.
Additionally, we compare LDR-GS and HDR-GS, both of which are vanilla 3DGS trained on post-processed LDR images and unprocessed RAW images, respectively.
The NeRF-based methods include RawNeRF~\cite{mildenhall2022nerf} and LDR-NeRF.
RawNeRF is a Mip-NeRF~\cite{barron2021mip} directly trained on noisy RAW images with weighted L2 loss and multi-exposure training strategy.
LDR-NeRF is a vanilla NeRF~\cite{mildenhall2020nerf} trained on the post-processed LDR images with L2 loss.

\setlength{\tabcolsep}{5pt}
\begin{table}[tb]
\centering
\vspace{-2mm}
\caption{
    Quantitative results on the test scenes of the RawNeRF~\cite{mildenhall2022nerf} dataset.
    The best result is in \best{bold} whereas the second best one is in \second{underlined}.
    TM indicates whether the tone-mapping function can be replaced for HDR rendering.
    For methods where the tone-mapping function can be replaced, 
    the metrics on sRGB are calculated using LDR tone-mapping for a fair comparison.
    The FPS measurement is conducted at a 2K (2016$\times$1512) resolution.
    Train denotes the training time of the method, measured in GPU$\times$H.
    \framework~achieves comparable performance with previous volumetric rendering based methods (RawNeRF~\cite{mildenhall2022nerf}), 
    but with 4000$\times$ faster rendering speed.
}
\label{tab:quant}
\begin{tabular}{ccccccccc}
\toprule
\multirow{2}{*}{Method} & \multirow{2}{*}{TM} & \multirow{2}{*}{FPS$\uparrow$} & \multirow{2}{*}{Train$\downarrow$} & \multicolumn{2}{c}{RAW}         & \multicolumn{3}{c}{sRGB}                           \\\cmidrule{5-9}
                        &                     &                                &                                    & PSNR$\uparrow$ & SSIM$\uparrow$ & PSNR$\uparrow$ & SSIM$\uparrow$ & LPIPS$\downarrow$\\\midrule
LDR-NeRF~\cite{mildenhall2020nerf}        & \ding{55}           & 0.007                          & 13.66                              & $-$            & $-$            & 20.0391        & 0.5541         & 0.5669            \\
LDR-3DGS~\cite{kerbl20233d}        & \ding{55}           & 153                            & 0.75                               & $-$            & $-$            & 20.2936        & 0.5477         & 0.5344            \\\midrule
HDR-3DGS~\cite{kerbl20233d}        & \ding{51}           & 238                            & 0.73                               & 56.4960        & 0.9926         & 20.3320        & 0.5286         & 0.6563            \\
RawNeRF~\cite{mildenhall2022nerf}         & \ding{51}           & 0.022                          & 129.54                             & 58.6920        & 0.9969         &\second{24.0836}& \best{0.6100}  & \best{0.4952}     \\
RawGS (Baseline)        & \ding{51}           & 176                            & 1.05                               &\second{59.2834}& \second{0.9971}& 23.3485        & 0.5843         & 0.5472            \\
\textbf{LE3D (Ours)}    & \ding{51}           & 103                            & 1.53                               & \best{61.0812} & \best{0.9983}  & \best{24.6984} & \second{0.6076}& \second{0.5071}   \\\bottomrule
\end{tabular}
\end{table}

\vspace{-1mm}
\paragraph{Quantitative evaluation}
\tabref{tab:quant} has shown the quantitative comparisons on the RawNeRF~\cite{mildenhall2022nerf} dataset.
Although NeRF-based methods have long training times and slow rendering speeds, they exhibit good metrics on sRGB.
This indicates that the volume rendering they rely on has strong noise resistance (mainly due to the dense sampling on each ray).
In contrast, 3DGS-based methods have inferior metrics compared to RawNeRF, due to their sparse scene representation and poor noise resistance.
Additionally, the splitting of gaussians depends on gradient strength, and supervision using noisy raw images affects this process, leading to incomplete structure recovery.
\framework~achieves better structure reconstruction suitable for downstream tasks through supervision on structure, depth distortion, and near-far regularizations, as detailed in~\secref{sec:applications}.
Note that the results of \framework~ are comparable to the previous volumetric rendering-based method, RawNeRF~\cite{mildenhall2022nerf}, in both quantitative and qualitative aspects.
However, it requires only 1\% of the training time and achieves a 3000$\times$-6000$\times$ rendering speed improvement.

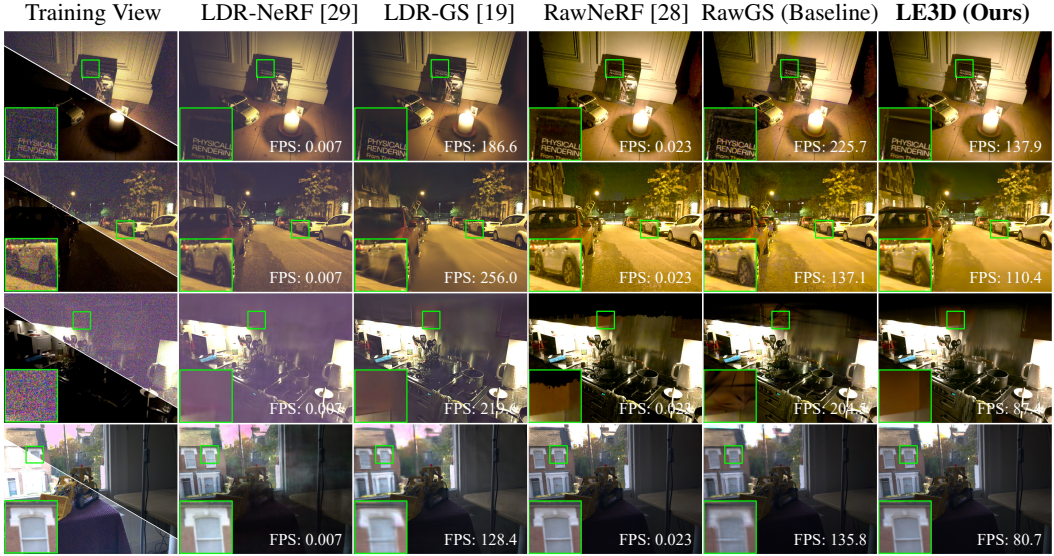
\begin{figure*}[ht]
    \centering
    \vspace{1mm}
    \begin{overpic}[width=\textwidth]{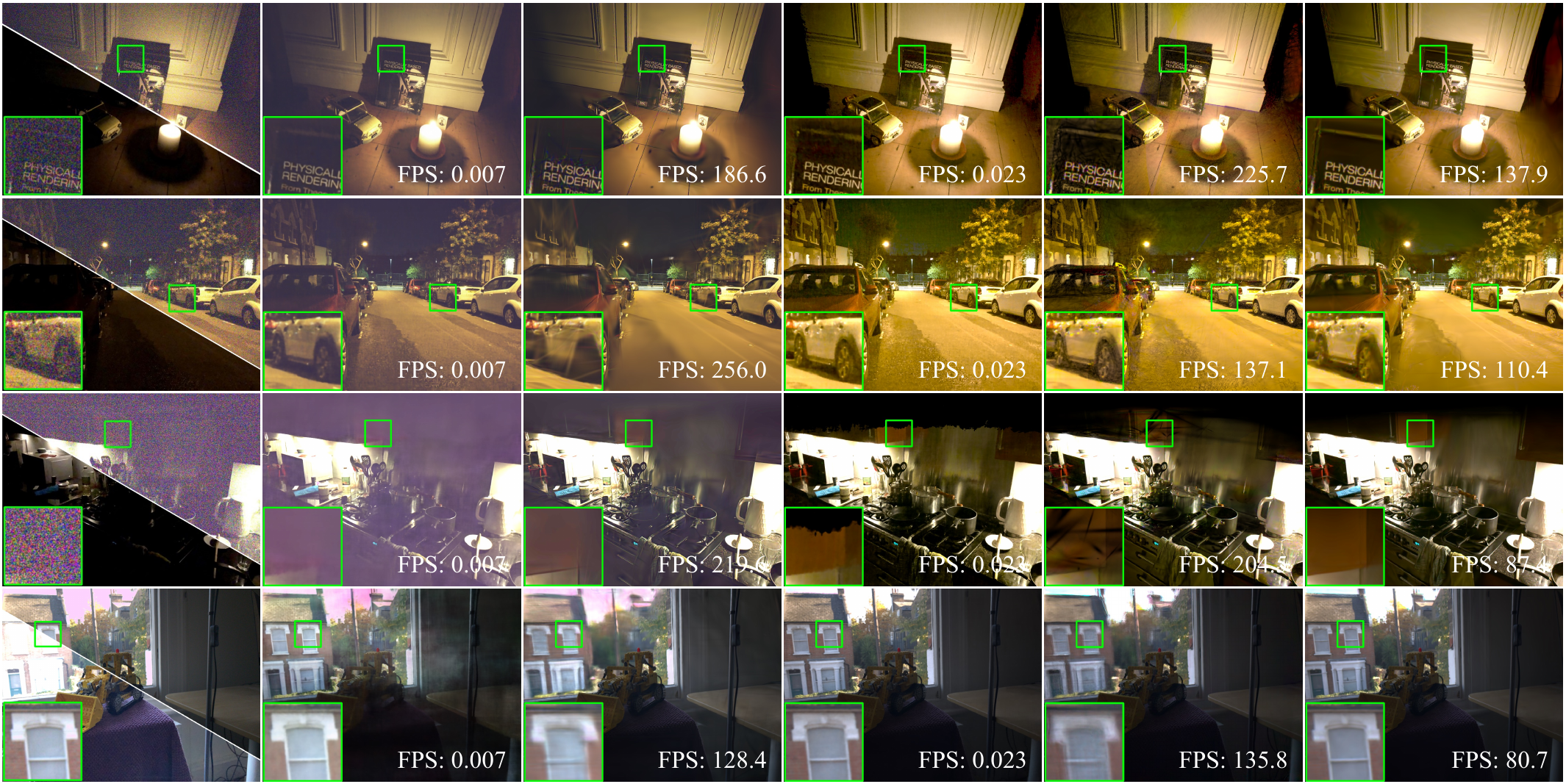}
        \put(2.15,51){\small{Training View}}
        \put(18.9,51){\small{LDR-NeRF~\cite{mildenhall2020nerf}}}
        \put(36.3,51){\small{LDR-GS~\cite{kerbl20233d}}}
        \put(51.5,51){\small{RawNeRF~\cite{mildenhall2022nerf}}}
        \put(66.5,51){\small{RawGS (Baseline)}}
        \put(85,51){\small{\textbf{\framework~(Ours)}}}
    \end{overpic}
    \vspace{-5mm}
    \caption{
        Visual comparison between \framework~and other reconstruction methods (\textit{Zoom-in for best view}).
        The training view contains two parts: the post-processed RAW image with linear brightness enhancement (up) and the image directly output by the device (down).
        By comparison to the 3DGS-based method, \framework~recovers sharper details in the distant scene and is more resistant to noise.
        Additionally, compared to NeRF-based methods, \framework~achieves comparable results with $3000\times$-$6000\times$ improvement in rendering speed.
    }
    \label{fig:scenes}
\end{figure*}

\vspace{-1mm}
\paragraph{Qualitative evaluation}
\figref{fig:scenes} has shown the qualitative comparisons on the RawNeRF~\cite{mildenhall2022nerf} dataset. 
We selected four scenes for comparison, including two indoor scenes and two outdoor scenes. 
The data for the first two scenes were collected with a single exposure, 
while the data for the latter two scenes included multiple exposures.
Compared to 3DGS~\cite{kerbl20233d}-based methods, \framework~demonstrates stronger noise resistance, particularly in the first two scenes.
Additionally, \framework~achieves better results in distant scene reconstruction. 
For example, in the second scene, \framework~produces a smoother sky compared to RawGS, 
and in the fourth scene, \framework~recovers distant details more sharply.
Compared to RawNeRF, \framework~typically produces smoother results while still effectively preserving details.
Most importantly, \framework~offers faster training times and rendering speeds.

\subsection{Ablation studies}

\begin{figure}
    \centering
    \begin{overpic}[width=\textwidth]{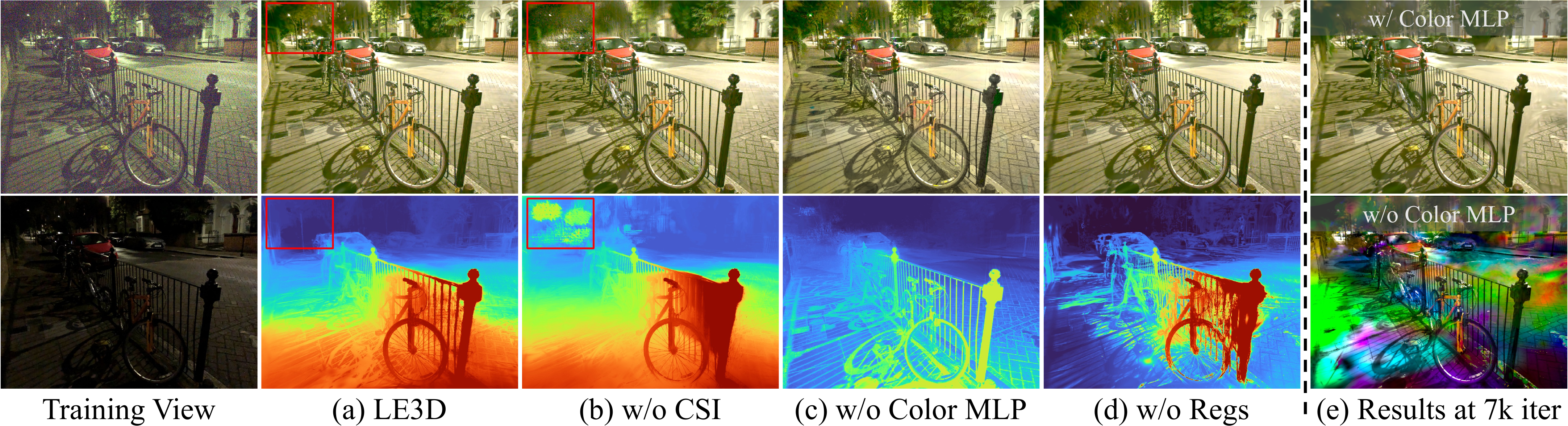}
    \end{overpic}
    \vspace{-6mm}
    \caption{
    Ablation studies on our purposed methods (\textit{Zoom-in for best view}). CSI in (b) and Regs in (d) denote Cone Scatter Initialization and Regularizations, respectively. (e) shows the rendering result of \framework~w/ or w/o Color MLP in early stages of training.
    }
    \vspace{-2mm}
    \label{fig:ablation}
\end{figure}

\paragraph{Cone Scatter Initialization (CSI)}
In low-light environments, COLMAP struggles to obtain a high-quality sparse point cloud. Although 3DGS demonstrates its robustness to the quality of the initial point cloud, it still encounters difficulties in achieving optimal geometric reconstruction within insufficient initialized areas.
It can be observed from \figref{fig:ablation} (b) that the methods without CSI tend to generate gaussians at incorrect depths and lack fine details. 
Conversely, CSI extends the depth coverage of the scene, enabling 3DGS to generate gaussians at relatively accurate depths and exhibit superior detail representation.
A comparative analysis between \figref{fig:ablation} (a) and \figref{fig:ablation} (b) suggests that our initialization technique plays a pivotal role in achieving accurate and detailed 3D reconstruction. 

\vspace{-2mm}
\paragraph{Color MLP}
\label{sec:ablation_colormlp}
Replacing SH with Color MLP not only enhances the expressiveness of our model but also introduces greater stability during the optimization process.
\figref{fig:ablation} (e) reveals that the method employing SH rather than Color MLP exhibits strange color representations early in the training phase, due to the inability of SH to adequately represent the RAW linear color space. Although the rendered image may appear similar to those produced by the \framework, the underlying issues have significantly affected the final structural reconstruction, as depicted in \figref{fig:ablation} (c).

\vspace{-2mm}
\paragraph{Regularizations}
\label{sec:ablation_reg}
Superior visual effects in 3D are contingent upon a robust 3D structure reconstruction, which in turn significantly enhances the performance of downstream tasks such as refocusing. 
To this end, we implement depth distortion regularization $\mathcal{R}_{dist}$ and near-far regularization $\mathcal{R}_{nf}$ to constrain the gaussians, ensuring their aggregation at the surfaces of objects and thereby improving the quality of structural reconstruction.
\figref{fig:ablation} (d) underscores the substantial enhancement our proposed regularizations provide in reconstructing the 3D structure of scenes.

\section{More applications}
\label{sec:applications}

\begin{figure*}[tb]
    \centering
    \begin{overpic}[width=0.93\textwidth]{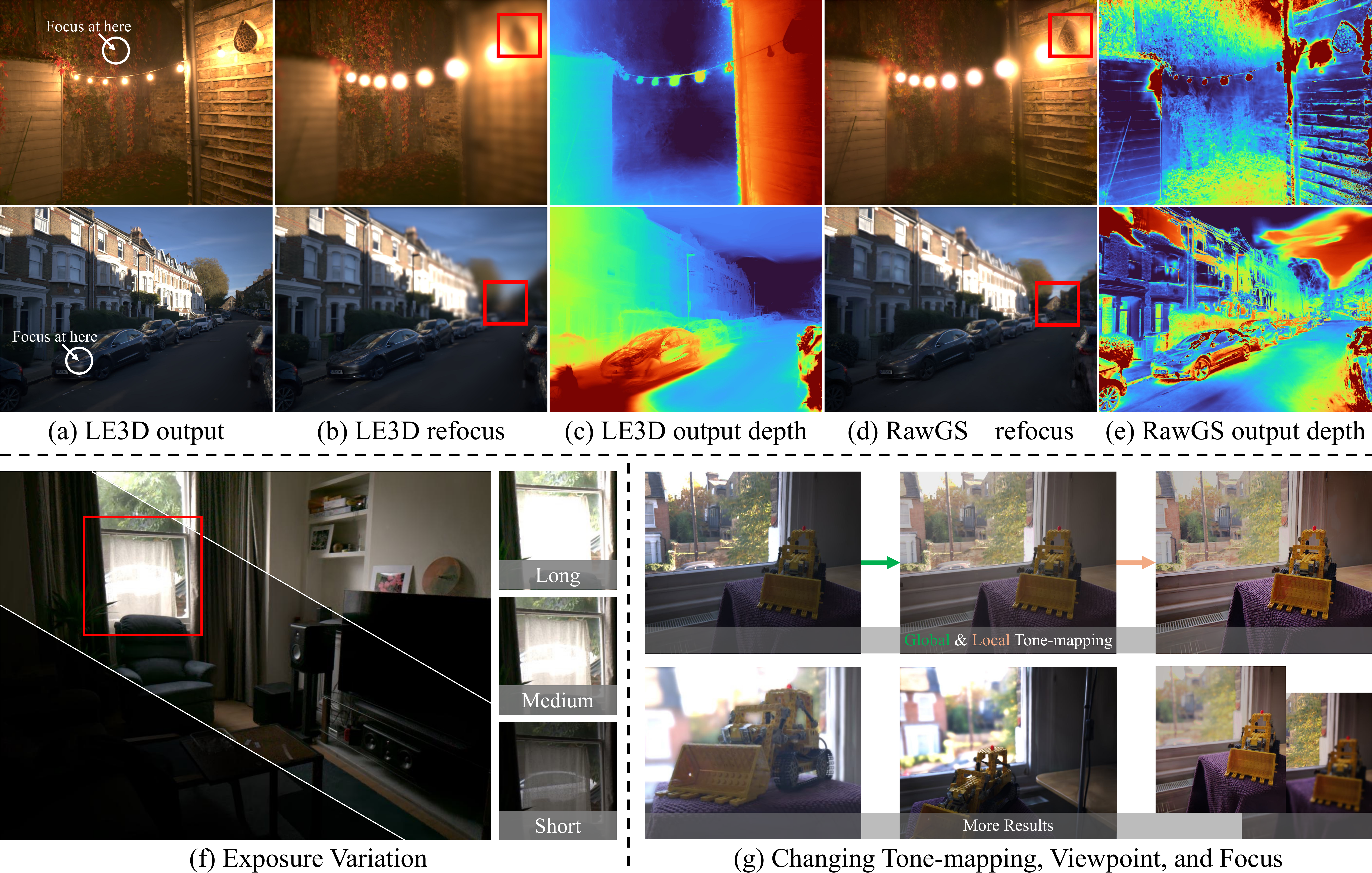}
        \put(70.7,31.6){\small{$\star$}}
    \end{overpic}
    \vspace{-3mm}
    \caption{
        \framework~supports various applications. RawGS$\star$ in (d) denotes using \framework's rendered image and RawGS's structure information as input for refocusing.
        (c, e) are the weighted depth rendered by \framework~and RawGS, respectively.
        (f) shows the same scene rendered by \framework~with different exposure settings.
        In (g), the ``{\textcolor{mygreen}{$\rightarrow$}}'' denotes global tone-mapping, 
        while the ``{\textcolor{myflesh}{$\rightarrow$}}'' represents local tone-mapping.
    }
    \label{fig:application}
    \vspace{-3mm}
\end{figure*}

\vspace{-1mm}
\paragraph{Refocus}
Structural information is crucial for tasks like refocusing. As discussed in~\secref{sec:ablation_reg}, \framework~benefits from the inclusion of depth distortion and near-far regularizations, 
which enhances its ability to learn structural details.
As shown in~\figref{fig:application} (b, d), \framework~achieves more realistic refocusing effects due to its superior structural information, as reflected in the depth shown in (c).
Conversely, RawGS suffers from foreground and background ambiguity in refocusing due to the lack of structural information.
Detailed refocusing algorithm will be released in the supplementary materials.

\vspace{-4mm}
\paragraph{Exposure variation and HDR tone-mapping}
\framework~can easily achieve exposure variation and recover details from overexposed data, as shown in~\figref{fig:application} (f).
\figref{fig:application} (g) showcases the various tone-mapping methods \framework~can implement, 
including global tone-mapping, such as color temperature and curve adjustments, and local tone-mapping using our re-implemented HDR+~\cite{hasinoff2016burst} 
(implementation details will be released in the supplementary materials).

Although RawNeRF~\cite{mildenhall2022nerf} can also perform similar applications, its inability to achieve \textbf{real-time} rendering significantly limits its use cases, such as real-time editing described in~\secref{sec:supp_viewer}.
\section{Conclusion}
\vspace{-3mm}

To address the long training times and slow rendering speeds of previous volumetric rendering-based methods,
we propose \framework~based on 3DGS.
Additionally, we introduce Cone Scatter Initialization and a tiny MLP for representing color in the linear color space.
This addresses the issue of missing distant points in nighttime scenes with COLMAP initialization. 
It also replaces spherical harmonics with the tiny color MLP, effectively representing the linear color space.
Finally, we enhance the structural reconstruction with the proposed depth distortion and near-far regularization, 
enabling more effective and realistic downstream tasks.
Benefiting from the rendering images in the linear color space, \framework~can achieve more realistic exposure variation and HDR tone-mapping in real-time, 
expanding the possibilities for subsequent HDR view synthesis processing.

{
\small
\bibliographystyle{plain}
\bibliography{egbib}
}


\appendix

\section{Implementation details and more ablation studies}
\label{sec:supp_sec_A}

\subsection{Affine alignment}
\label{sec:supp_affine_alignment}

Since all training views in the RawNeRF~\cite{mildenhall2022nerf} dataset are captured with a fast shutter, while the test views (ground truth) are captured with a slow shutter, 
linear enhancement is needed during testing for alignment. 
However, due to color bias (non-zero-mean noise for high ISO, ~\cite{wei2021physics}), direct linear enhancement does not achieve perfect alignment. 
Therefore, affine alignment is performed on both the output and ground truth during testing. 
In RawNeRF~\cite{mildenhall2022nerf}, this process is done as the following procedure:
\begin{equation}
    a=\frac{\overline{xy}- \overline{x}\overline{y}}{\overline{x^2}-\overline{x}^2}=\frac{\mathrm{Cov}{(x, y)}}{\mathrm{Var}(x)}, b=\overline{y}-a\overline{x}.
\end{equation}
where $\overline{x}$ is the mean of $x$. And $x, y$ are the groundtruth and the final output, respectively.
This is the least-squares fit of an affine transform $ax+b\approx y$. 
At test time, we first process $y$ with $(y - b) / a$, then calculate the metric.
For those methods whose output is in RAW linear color space, the affine alignment process is only done \textit{once} in the RAW color space.
While for other methods (LDR-NeRF~\cite{mildenhall2020nerf}, LDR-3DGS~\cite{kerbl20233d}) which can only output in RGB color space, the affine alignment process is done in the RGB color space.

\subsection{More ablation studies}
\label{sec:supp_more_ablation}

\begin{figure*}[t]
    \centering
    \begin{overpic}[width=\textwidth]{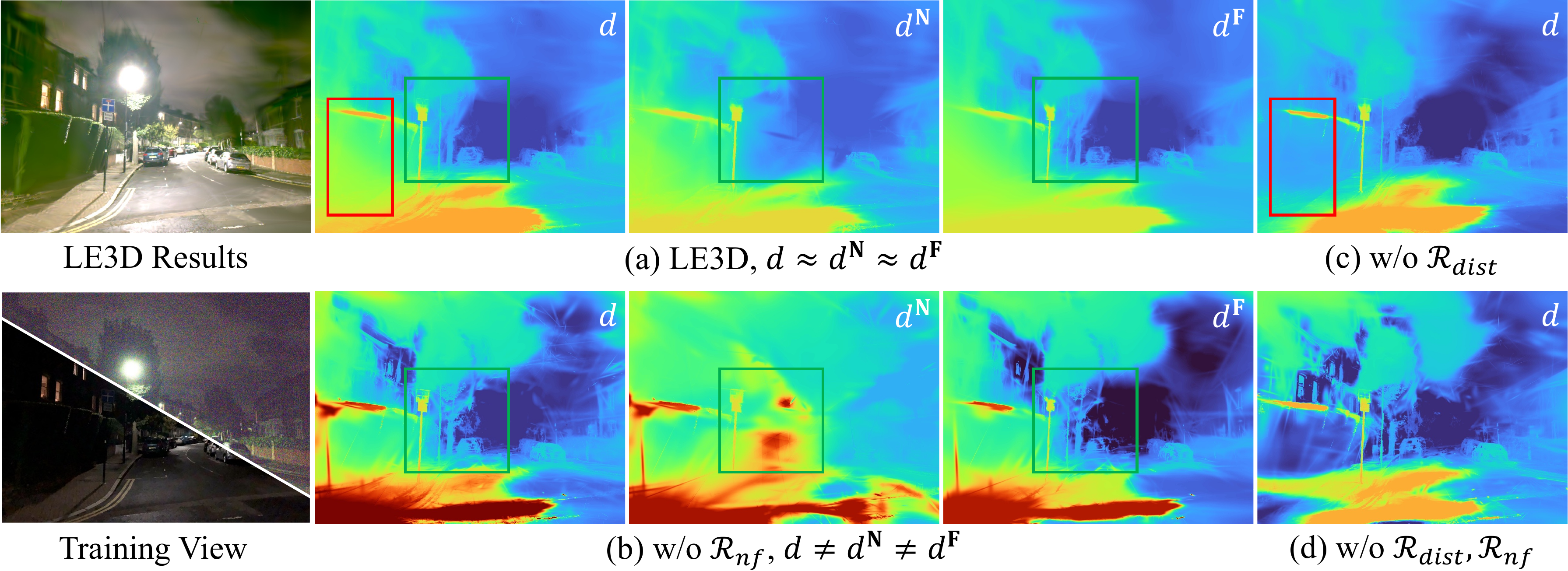}
    \end{overpic}
    \caption{
        Ablation on each of the regularization. Since both $\mathcal{R}_{dist}$ and $\mathcal{R}_{nf}$ are regularization terms intended to strengthen the structural representation, we have elected to display only the depth map for the sake of clarity. In addition, to demonstrate the effect of $\mathcal{R}_{nf}$ in aligning $d, d^\mathbf{N}, d^\mathbf{F}$, we have also visualized $d^\mathbf{N}$ and $d^\mathbf{F}$ as mentioned in \secref{sec:method_regs}.
    }
    \label{fig:supp_regs_ablation}
\end{figure*}

\paragraph{Ablation on each of the regularizations}
\figref{fig:supp_regs_ablation} (a,b) presents the visualization of $d^\mathbf{N}$ and $d^\mathbf{F}$ adjacent to the depth map. In an ideal scenario, both $d^\mathbf{N}$ and $d^\mathbf{F}$ should align with $d$, ensuring that the weights along each ray are concentrated at surface. The comparison between \figref{fig:supp_regs_ablation} (a,b) demonstrates that the incorporation of near-far regularization indeed encourages $d^\mathbf{N}$ and $d^\mathbf{F}$ to progressively align with $d$. This alignment results in a more refined representation of the three-dimensional structure, capturing better details of the scene's geometry.
The comparison between \figref{fig:supp_regs_ablation} (a,c) elucidates the adverse effects of omitting distortion regularization. Without such constraints, the model would fail to produce depth maps with a natural depth progression, with artifacts such as abrupt depth discontinuities or voids on planes. Such anomalies are indicative of significant issues in the reconstruction of the scene's geometry.

\begin{figure*}[t]
    \centering
    \begin{overpic}[width=\textwidth]{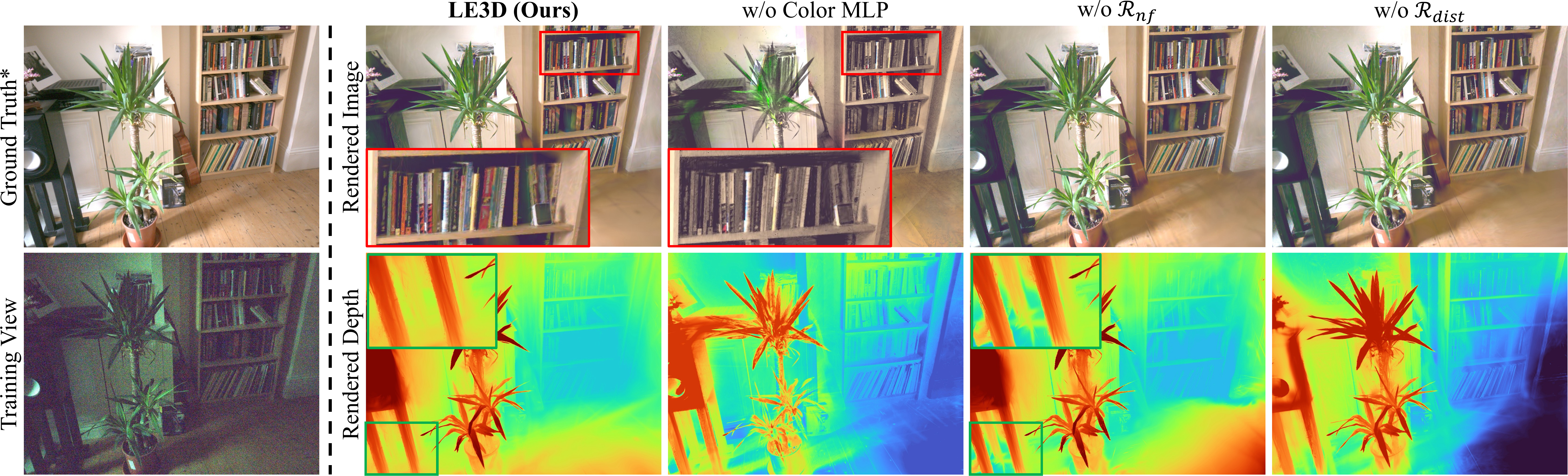}
    \end{overpic}
    \caption{
        Visualization results for ablation studies on the test scene.
        The Ground Truth\* denotes the raw image averaged from a burst set with slow shutter to perform denoising.
        It can be observed that the results of w/o Color MLP show significant color degradation, while the results of w/o $\mathcal{R}{nf}$ and w/o $\mathcal{R}{dist}$ exhibit structural degradation.
    }
    \label{fig:supp_quant_ablation}
\end{figure*}

\paragraph{Ablation studies on test scenes}
As shown in~\tabref{tab:supp_quant_ablation}, \framework~has shown superior performance over all ablated methods. 
The results without Color MLP are the worst because the SH used in vanilla 3DGS is not suitable for representing colors in the RAW linear color space.
As shown in~\figref{fig:supp_quant_ablation}, the results without Color MLP are noticeably desaturated and appear gray.
Both w/o $\mathcal{R}_{nf}$ and w/o $\mathcal{R}_{dist}$ show degraded depth, as seen in~\figref{fig:supp_quant_ablation}.
Additionally, it can be observed that w/o Color MLP also has poor structural information. 
This is mainly due to instability during the early stages of training, leading to suboptimal depth map reconstruction, as discussed in~\secref{sec:ablation_colormlp}.

\paragraph{The stability of~\framework}

\begin{figure*}[t]
    \centering
    \begin{overpic}[width=\textwidth]{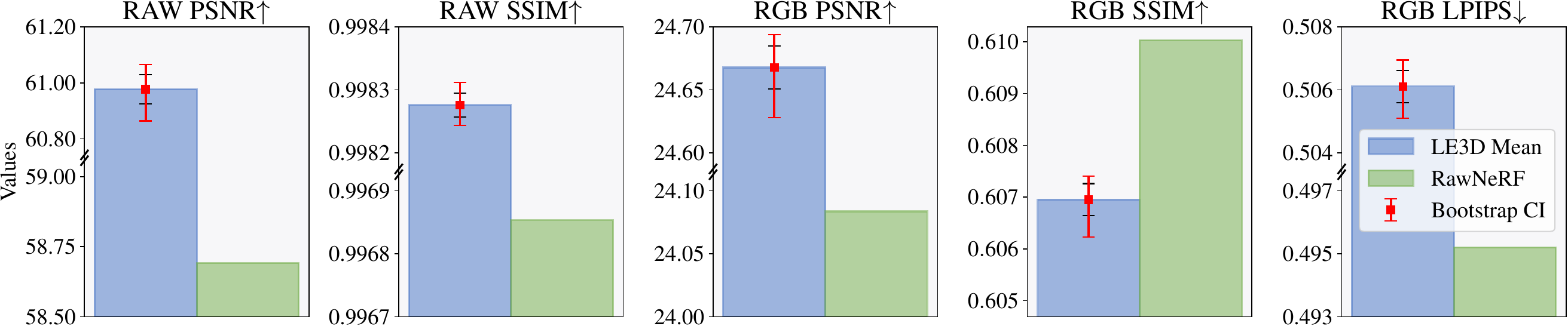}
    \end{overpic}
    \caption{
        The error bars of our proposed \framework~and comparison with RawNeRF~\cite{mildenhall2022nerf}.
    }
    \label{fig:supp_error_bar}
\end{figure*}

Due to the random initialization of our Color MLP, we trained 9 versions of \framework~using 9 different random seeds to test the stability of \framework. 
We then compared their metrics on the test set, as shown in~\figref{fig:supp_error_bar}.
It can be observed that the stability of \framework~is remarkably high, and the overall fluctuations do not impact the experimental conclusions.

\setlength{\tabcolsep}{8pt}
\begin{table}[tb]
\centering
\caption{
    Quantitative results of the ablation studies.
    Notice that, since Cone Scatter Initialization (CSI) is used to supplement the point cloud in distant scenes, 
    and the test scenes do not contain distant views (all being indoor scenes), \framework~\textit{does not apply CSI} in this context.
    The ablation study of CSI can be found in~\figref{fig:ablation} (a, b), which shows significant differences in distant view.
    Best results is denoted in {\textbf{bold}}.
    The rank is indicated in the lower right corner of each metrics.
}
\label{tab:supp_quant_ablation}
\begin{tabular}{cccccc}
\toprule
\multirow{2}{*}{Method} & \multicolumn{2}{c}{RAW}         & \multicolumn{3}{c}{sRGB}                           \\\cmidrule{2-6}
                        & PSNR$\uparrow$ & SSIM$\uparrow$ & PSNR$\uparrow$ & SSIM$\uparrow$ & LPIPS$\downarrow$\\\midrule
w/o Color MLP           & 59.4483$_{(\textbf{4})}$        & 0.9969$_{(\textbf{4})}$         & 23.1884$_{(\textbf{4})}$        & 0.5862$_{(\textbf{4})}$         & 0.5635$_{(\textbf{4})}$           \\
w/o $\mathcal{R}_{dist}$& 60.5202$_{(\textbf{3})}$        & 0.9981$_{(\textbf{3})}$         & 24.3615$_{(\textbf{3})}$        & 0.6007$_{(\textbf{3})}$         & 0.5087$_{(\textbf{2})}$           \\
w/o $\mathcal{R}_{nf}$  & 60.7144$_{(\textbf{2})}$        & 0.9982$_{(\textbf{2})}$         & 24.5705$_{(\textbf{2})}$        & 0.6043$_{(\textbf{2})}$         & 0.5096$_{(\textbf{3})}$           \\\midrule
\textbf{LE3D (Ours)}    & \textbf{61.0812}$_{(\textbf{1})}$ & \textbf{0.9983}$_{(\textbf{1})}$ & \textbf{24.6984}$_{(\textbf{1})}$ & \textbf{0.6077}$_{(\textbf{1})}$ & \textbf{0.5071}$_{(\textbf{1})}$\\\bottomrule
\end{tabular}
\end{table}

\section{Interactive viewer}
\label{sec:supp_viewer}

\begin{figure*}[t]
    \centering
    \begin{overpic}[width=\textwidth]{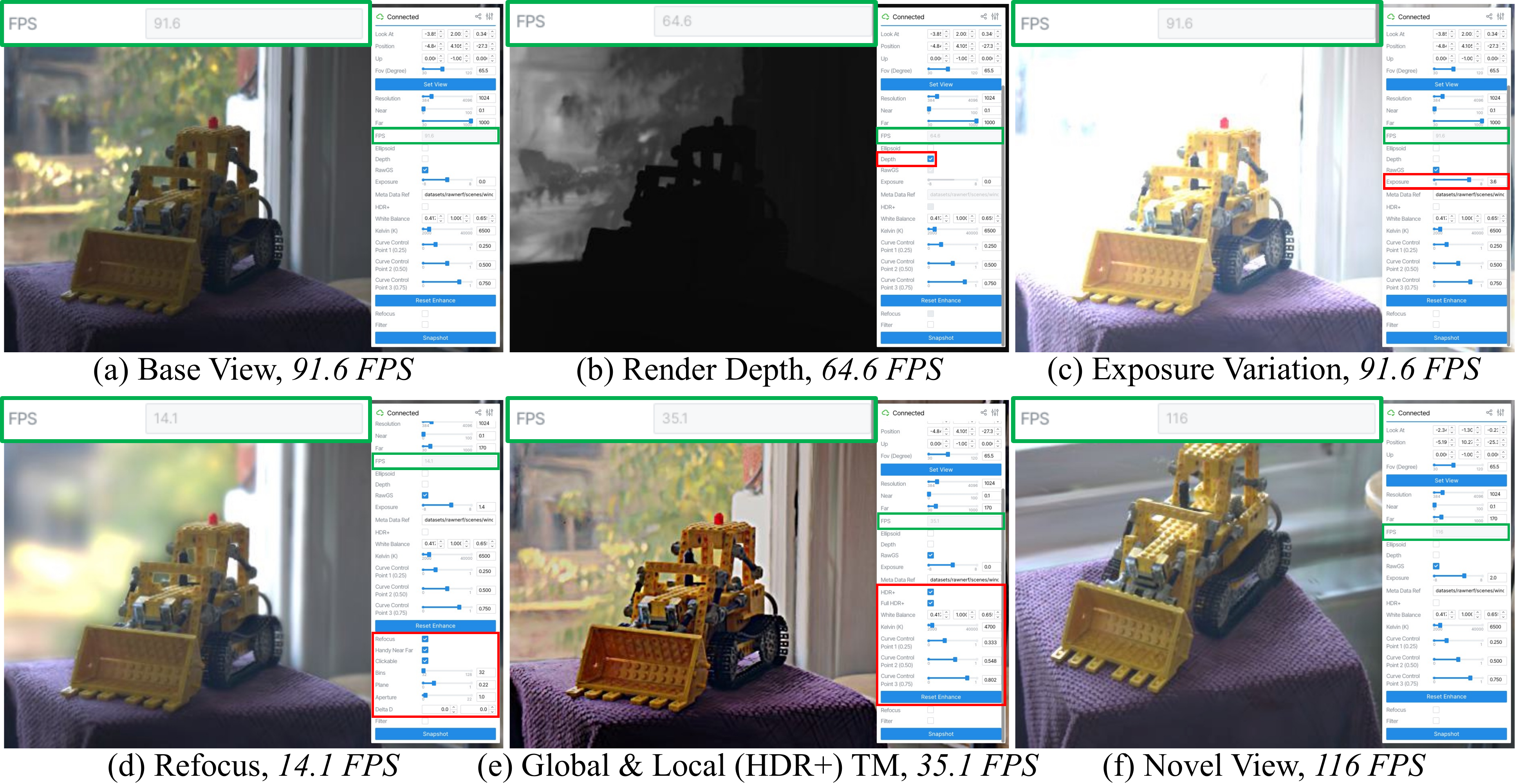}
    \end{overpic}
    \caption{
        Some screenshots of our interactive viewer, which can perform (b) depth rendering, 
        (c) exposure variation, (d) refocus, (e) global \& local tone-mapping and (f) novel view rendering.
        The FPS are emphasized by the \textcolor{mygreen}{green} bounding box, and the changed rendering parameters are emphasized in \textcolor{myred}{red} bounding box.
    }
    \label{fig:supp_viewer}
\end{figure*}

\figref{fig:supp_viewer} has shown some screenshots and downstream tasks results of our interactive viewer which is built upon Viser~\cite{nerfstudio2023viser}.
Besides refocusing, most of the downstream tasks can be performed in real-time.
While for refocusing, most of the time is spent on the gaussian blur according to the refocusing algorithm (due to the large gaussian blur kernel size) rather than on rendering.

\section{More results}

\paragraph{Detailed comparisons between 3DGS-based methods and \framework}

\begin{figure*}[t]
    \centering
    \vspace{1mm}
    \begin{overpic}[width=\textwidth]{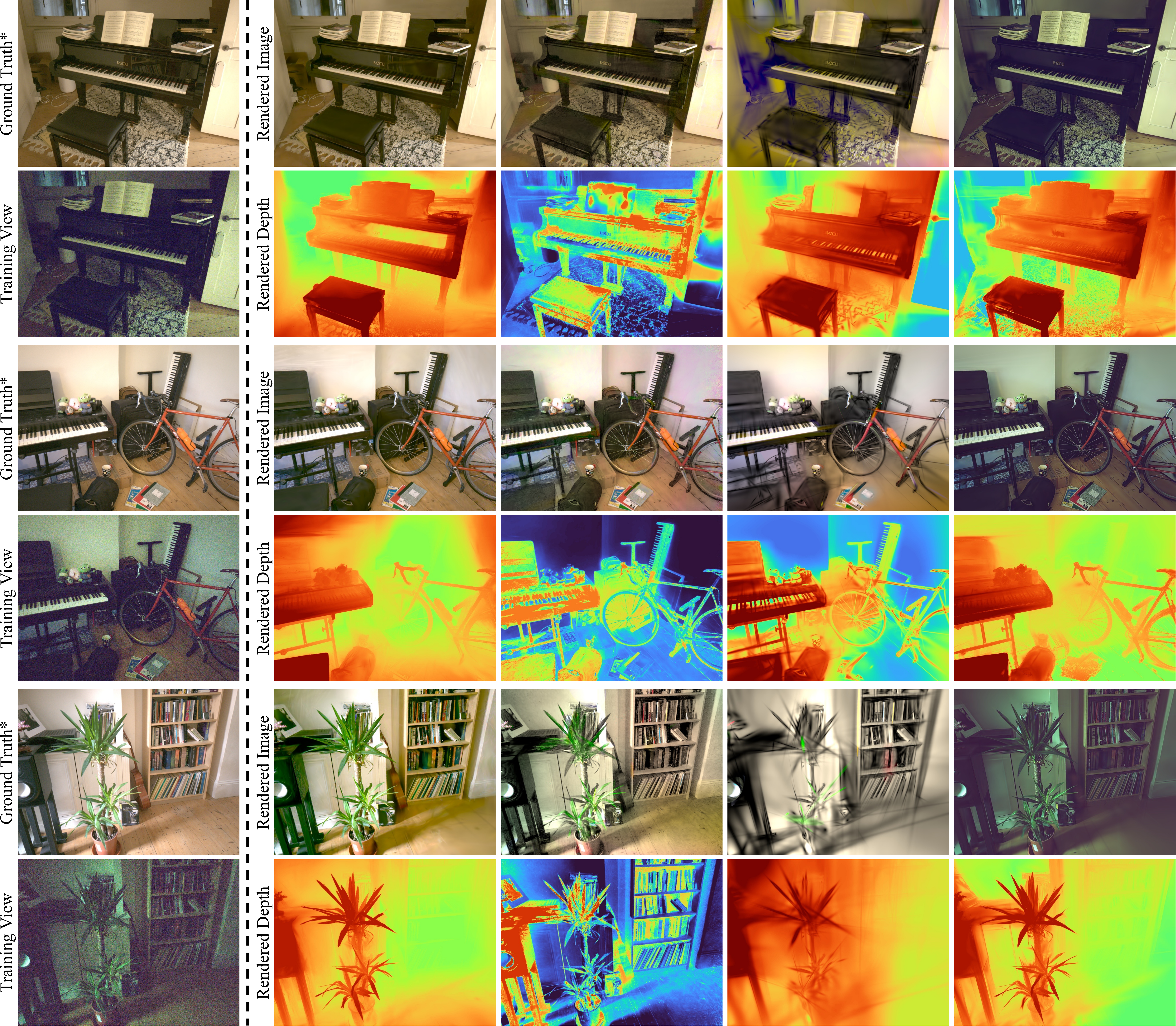}
        \put(28, 87.7){\scriptsize{\textbf{\framework~(Ours)}}}
        \put(46, 87.7){\scriptsize{RawGS (Baseline)}}
        \put(67, 87.7){\scriptsize{HDR-GS~\cite{kerbl20233d}}}
        \put(86.7, 87.7){\scriptsize{LDR-GS~\cite{kerbl20233d}}}
    \end{overpic}
    \caption{
        Comparison between \framework~and other 3DGS-based methods (\textit{Zoom-in for best view}).
        All the results are the direct output of each model, not being applied by affine alignment.
        The Ground Truth\* denotes the raw image averaged from a burst set with slow shutter to perform denoising.
    }
    \label{fig:supp_gs_compare}
\end{figure*}

As shown in~\figref{fig:supp_gs_compare}, \framework~achieves better structure reconstruction than our baseline RawGS (3DGS trained with RawNeRF's loss and multiple exposure training strategy). 
And compare with LDR-GS (trained on the LDR images) and HDR-GS (trained directly on the RAW data), \framework~achieves better color reconstruction results as well as perform better denoising ability.
We also found that LDR-GS and HDR-GS have fewer reconstructed gaussians, resulting in faster rendering speeds but poor overall reconstruction quality.
Additionally, LDR-GS, trained on linear brightened LDR images, shows weaker resistance to color bias~\cite{wei2021physics}, resulting in severe color shifts in the final output.
We also found that the generally low values of RAW images lead to insufficient gradients, reducing the number of gaussian spliting. RawNeRF's weighted L2 loss (\eqnref{eq:rawnerf_pri}) strengthens supervision in dark areas but at the cost of structural information.
\framework~incorporates both the weighted L2 loss and depth distortion and near-far regularizations to constrain the structure, ultimately achieving the best structural and visual results.

\paragraph{More qualitative results}

\begin{figure*}[t]
    \centering
    \vspace{1mm}
    \begin{overpic}[width=\textwidth]{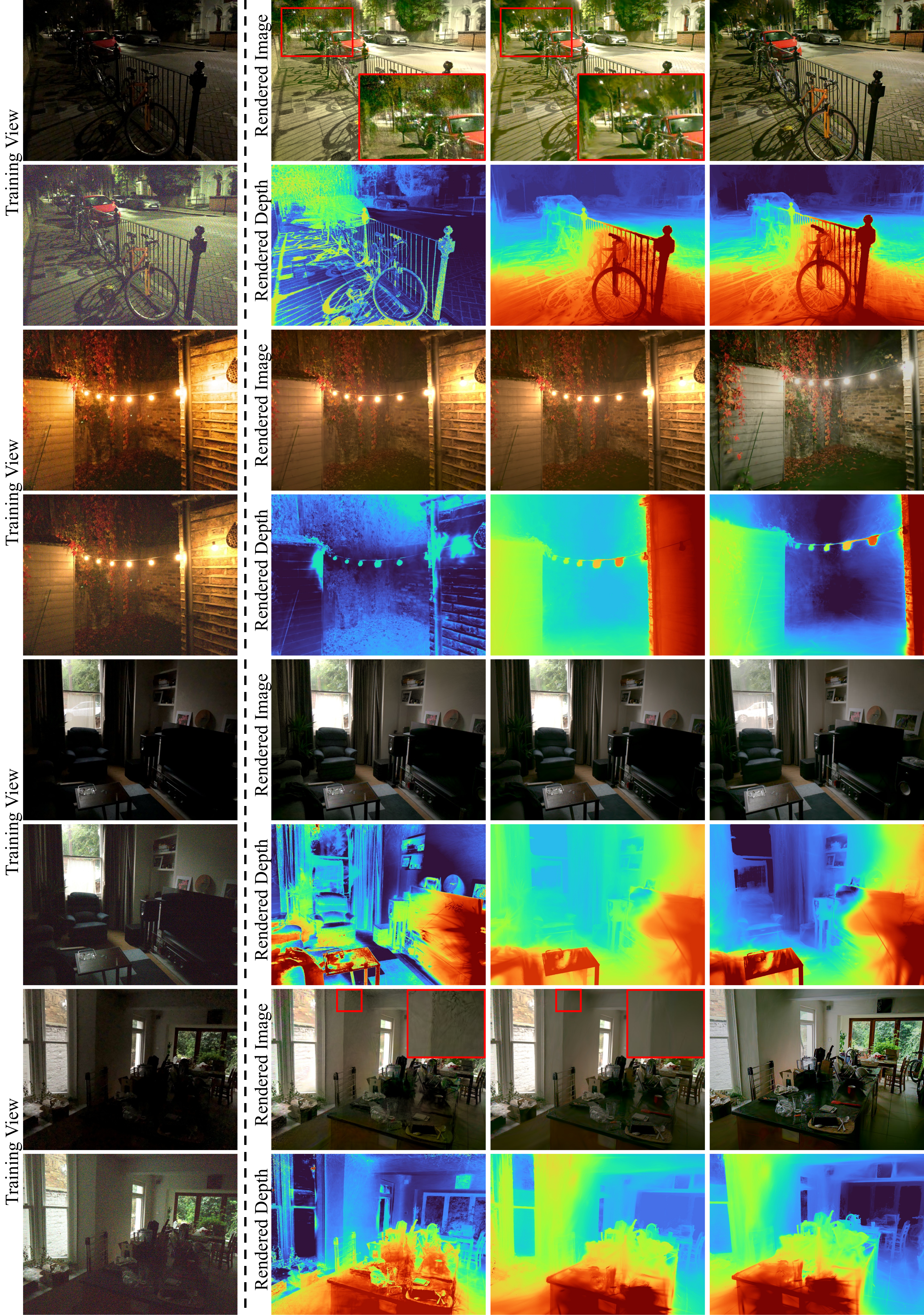}
        \put(22.8, 100.5){\small{RawGS (Baseline)}}
        \put(40.9, 100.5){\small{\textbf{\framework~(Ours)}}}
        \put(54.7, 100.5){\small{\textbf{\framework}, \scriptsize{Novel View, Edited}}}
    \end{overpic}
    \caption{
        Comparison between \framework~and RawGS (baseline, 3DGS trained with weighted L2 loss in~\eqnref{eq:rawnerf_pri} and multiple exposure strategy).
        It can be observed that \framework exhibits stronger noise resistance and color representation in low-light scenes.
        Additionally, it produces smoother and more accurate depth maps across all scenes.
    }
    \label{fig:supp_more_compare_1}
\end{figure*}

\begin{figure*}[t]
    \centering
    \vspace{1mm}
    \begin{overpic}[width=\textwidth]{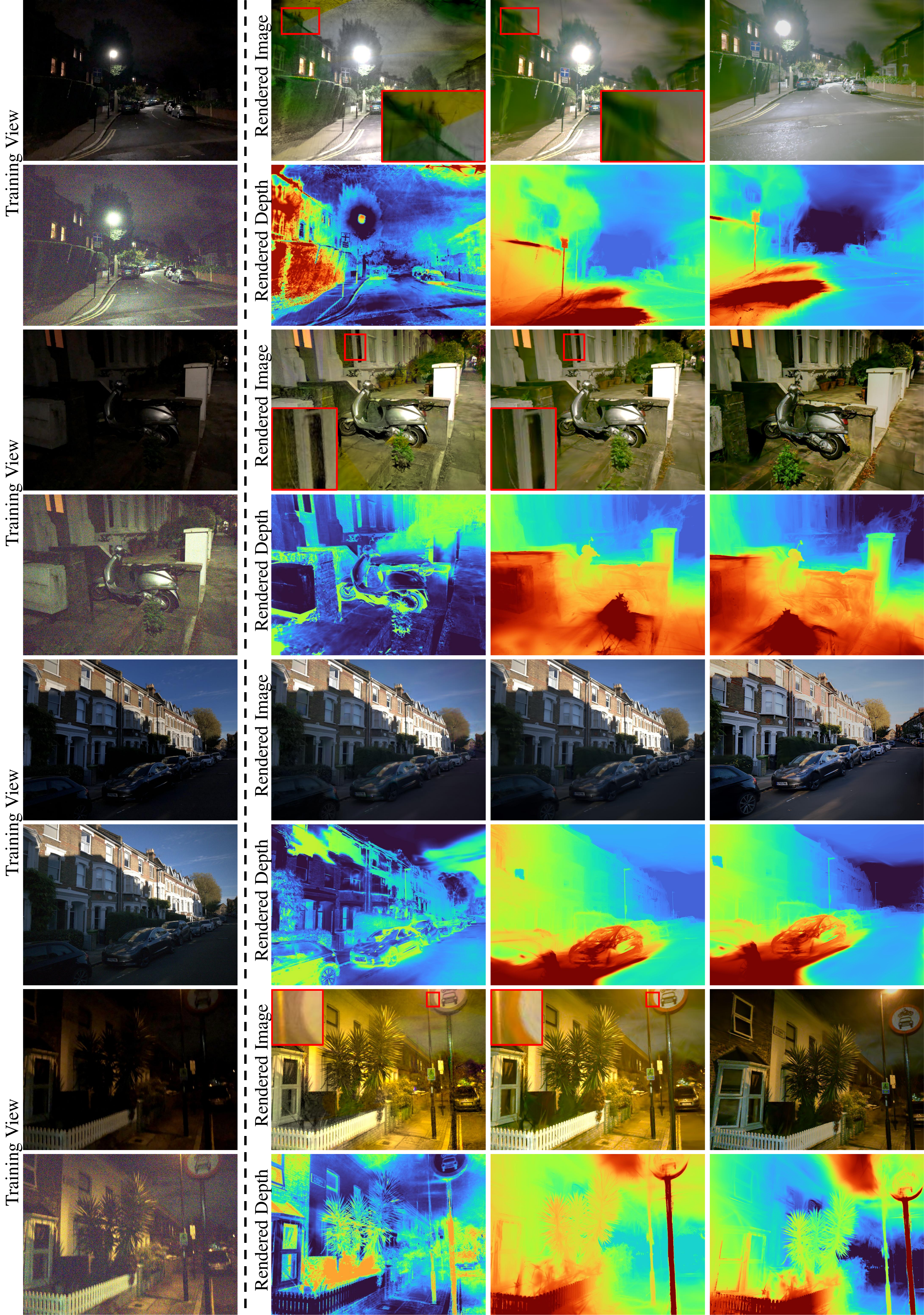}
        \put(22.8, 100.5){\small{RawGS (Baseline)}}
        \put(40.9, 100.5){\small{\textbf{\framework~(Ours)}}}
        \put(54.7, 100.5){\small{\textbf{\framework}, \scriptsize{Novel View, Edited}}}
    \end{overpic}
    \caption{
        Comparison between \framework~and RawGS (baseline, 3DGS trained with weighted L2 loss in~\eqnref{eq:rawnerf_pri} and multiple exposure strategy).
        It can be observed that \framework exhibits stronger noise resistance and color representation in low-light scenes.
        Additionally, it produces smoother and more accurate depth maps across all scenes.
    }
    \label{fig:supp_more_compare_2}
\end{figure*}

\figref{fig:supp_quant_ablation}, \figref{fig:supp_gs_compare}, \figref{fig:supp_more_compare_1}, \figref{fig:supp_more_compare_2}
has shown more qualitative comparisons between \framework~and 3DGS~\cite{kerbl20233d}-based methods.
From the figures above, it is evident that \framework~demonstrates superior noise resistance and color representation capabilities.
Additionally, \framework~produces smoother and more accurate depth maps, which are essential for downstream tasks like refocusing.
It worth noting that volumetric rendering-based methods, such as RawNeRF~\cite{mildenhall2022nerf}, cannot achieve real-time rendering, which significantly limits their applications (including real-time scene editing).
Therefore, we do not compare them here. For comparisons between \framework~and volumetric rendering based methods, please refer to~\tabref{tab:quant} and~\figref{fig:scenes}.

\end{document}